\documentclass[runningheads]{llncs}
\usepackage{graphicx}
\usepackage{comment}
\usepackage{amsmath,amssymb} 
\usepackage{color}
\usepackage{mathrsfs}

\usepackage{booktabs}
\usepackage{color}

\begin{document}
\pagestyle{headings}
\mainmatter
\def\ECCVSubNumber{5859}  

\title{Hybrid Models for Open Set Recognition} 


 \author{Hongjie Zhang\inst{1} \and
 Ang Li\inst{2} \and
 Jie Guo\inst{1}\and
 Yanwen Guo\inst{1}$^*$}
 \authorrunning{H. Zhang et al.}
 \institute{State Key Laboratory for Novel Software Technology, Nanjing University, Nanjing 210023, China\\
 \email{hjzhang@smail.nju.edu.cn, \{guojie, ywguo\}@nju.edu.cn}
 \and
 DeepMind, Mountain View, CA, USA \\
 \email{anglili@google.com}}

\maketitle
\makeatletter{\renewcommand*{\@makefnmark}{}
\footnotetext{$^*$Corresponding Author.}\makeatother}

\begin{abstract}
Open set recognition requires a classifier to detect samples not belonging to any of the classes in its training set. Existing methods fit a probability distribution to the training samples on their embedding space and detect outliers according to this distribution. The embedding space is often obtained from a discriminative classifier. However, such discriminative representation focuses only on known classes, which may not be critical for distinguishing the unknown classes. We argue that the representation space should be jointly learned from the inlier classifier and the density estimator (served as an outlier detector). We propose the OpenHybrid framework, which is composed of an encoder to encode the input data into a joint embedding space, a classifier to classify samples to inlier classes, and a flow-based density estimator to detect whether a sample belongs to the unknown category. A typical problem of existing flow-based models is that they may assign a higher likelihood to outliers. However, we empirically observe that such an issue does not occur in our experiments when learning a joint representation for discriminative and generative components. Experiments on standard open set benchmarks also reveal that an end-to-end trained OpenHybrid model significantly outperforms state-of-the-art methods and flow-based baselines.

\end{abstract}

\section{Introduction}
\begin{figure}[!tb]
\centering
	\includegraphics[width=1.0\textwidth]{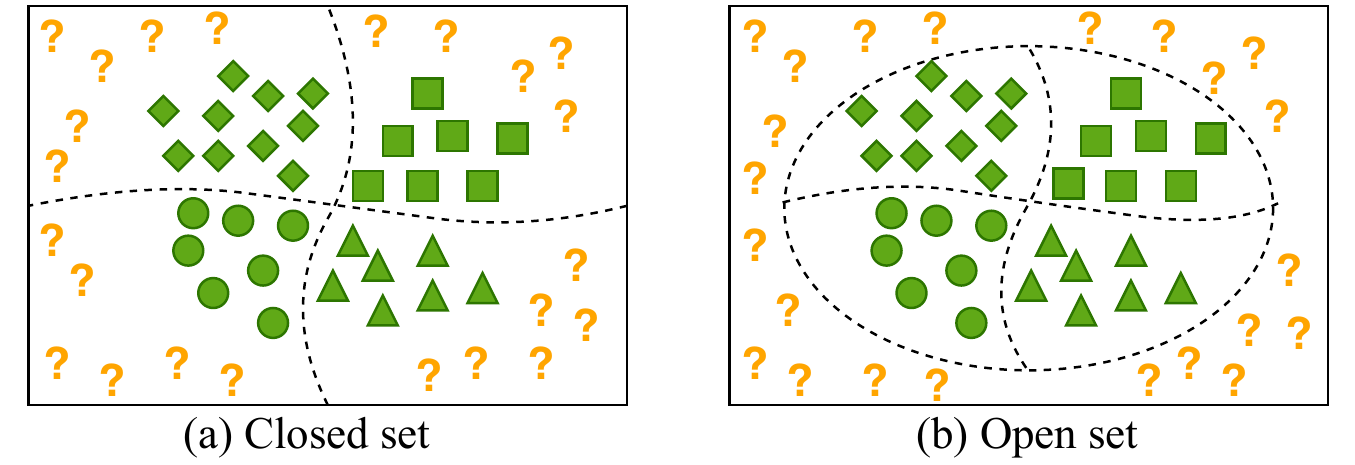}
 	\caption{Decision boundaries of a closed set classifier (a) and an open set classifier (b). Green symbols indicate known samples (different shapes represent different classes), and orange question marks indicate unknown samples. Dashed lines indicate the decision boundaries. (a) Closed set leads to unbounded decision boundaries of a typical 4-class classifier. Unknown samples are forced to be classified into one of known classes. (b) open set results in bounded decision boundaries for a 5-class classifier, which can classify both known and unknown samples.}
	\label{pullfigure}
\end{figure}

Image classification is a core problem in computer vision. However, most of the existing research is based on the closed-set assumption, \textit{i.e.}, training set is assumed to cover all classes that appear in the test set. This is an unrealistic assumption. Even with a large-scale image dataset, such as ImageNet \cite{krizhevsky2012imagenet}, it is impossible to cover all scenarios in the real world. When a closed-set model encounters an out-of-distribution sample, it is forced to identify it as a known class, which can cause issues in many real-world applications. We instead study the ``open-set'' problem where the test set is assumed to contain both known and unknown classes. So the model has to classify samples into either known (inlier) classes or the unknown (outlier) category. Figure \ref{pullfigure} illustrates the difference of classification decision boundaries under open set and closed set assumptions.

Identifying unknown samples is naturally challenging because they are not observed during training. 
Existing approaches fit a probability distribution of the training samples at their embedding space, and detect unknown samples according to such distribution. Since the feature representation of unknown classes is unknown, most of the methods operate on a discriminative feature space obtained from a supervised classifier trained on known classes. A thresholding on this probability distribution is then used to detect samples from unknown classes. A common approach along this direction is to threshold on SoftMax responses, but \cite{bendale2016towards} has conducted experiments to show that it reachs only sub-optimal solutions to open set recognition. Some variants have been proposed to better utilize the SoftMax scores \cite{ge2017generative,neal2018open,shu2017doc}. These methods modify the SoftMax scores to perform both unknown detection while maintaining its classification accuracy. It is extremely challenging to find a single score measure on the SoftMax layer, that can perform well on both the generative and discriminative tasks. 
We believe the discriminative feature space learned by classification of inlier classes may not be sufficiently effective for identifying outlier classes. So we propose to employ a flow-based generative model for outlier detection, and learn a joint feature space in an end-to-end manner from both the classifier and the density estimator.

Flow-based models have recently emerged \cite{behrmann2018invertible,chen2019residual,dinh2014nice,dinh2016density,kingma2018glow}, allowing a neural network to be invertible. They can fit a probability distribution to training samples in an unsupervised manner via maximum likelihood estimation. The flow models can predict the probability density of each example. When the probability density of an input sample is large, it is likely to be part of the training distribution (known classes). And the outlier samples (unknown class) usually have a small probability density value. The advantage of flow-based models is that they do not require the intervention of a classifier when fitting a probability distribution, and one can directly apply a thresholding model on these probability values without modifying the scores of any known classes. 

Flow-based models have been adopted to solve out-of-distribution detection \cite{nalisnick2019hybrid,nalisnick2018deep,hendrycks2018deep}, but have not yet been considered the open set recognition problem. Most related to our approach, \cite{nalisnick2019hybrid} proposed a deep invertible generalized linear model (DIGLM), which is comprised of a generalized linear model (GLM) stacked on top of flow-based model. They use the model's natural rejection rule based on the probability generated by flow-based model to detect unknown inputs, and directly classify known samples with the features used to fit the probability distribution. 
Our work differs in that instead of adding a classifier on top of flow model's embedding, we propose to learn a joint embedding for both the flow model and the classifier. Our insight is that the embedding space learned from only flow-based model may not have sufficient discriminative expressiveness. 

We empirically observe in our experiments that learning a joint embedding space resolves a common issue in flow-based model that the flow-based model may assign higher likelihood to OOD inputs (mentioned in \cite{hendrycks2018deep,ren2019likelihood,nalisnick2018deep}). 
This issue was considered in  \cite{kamoi2019likelihood}, the underlying factor of which is believed to be to the inconsistency between a uni-modal prior distribution and a multi-modal data distribution. In our framework, the deep network can well represent the multi-modal distribution of the input data, which is probably the reason for the improved performance of flow models.

We perform extensive experiments on various benchmarks including MNIST, SVHN, CIFAR10 and TinyImageNet. The proposed OpenHybrid model outperforms both state-of-the-art methods \cite{bendale2016towards,ge2017generative,neal2018open,oza2019c2ae,yoshihashi2019classification} and hybrid model baselines \cite{nalisnick2019hybrid,hendrycks2018deep} in these benchmarks. We further compare our method with an additional baseline which uses a pre-trained encoder and the result suggests the importance of jointly training both the classifier and the flow-based model.

\subsubsection{Contribution.} The contribution of this paper can be summarized as follows:
\begin{enumerate}
\item To the best of our knowledge, we are the first to incorporate a generative flow-based model with a discriminative classifier to address open set recognition, while most of the existing open set approaches focus on either using the softmax logits or adversarial training.
\item We propose the OpenHybrid model that learns a joint representation between the classifier and flow density estimator. Our approach ensures that the inlier classification is unaffected by outlier detection. We find joint training an important contributing factor, according to the ablation study.
\item 
A known issue of flow-based models is that they may assign higher likelihood to unknown inputs. However, we do not observe such phenomenon in OpenHybrid, possibly because our encoder fits the multi-modal input distribution to a latent space suitable to the unimodal assumption of flow models.
\item We conducted extensive experiments on various open set image classification datasets and compared our approach against  state-of-the-art open set methods and flow-based baseline models. Our approach achieves significant improvement over these baseline methods.
\end{enumerate}

\section{Related Work}

\subsection{Open Set Recognition}
Open set recognition has been surprisingly overlooked, though it has more practical value than the common closed set setting. Existing methods on this topic can be broadly classified into two categories: discriminative and generative models. 

\textit{Discriminative methods.} Before the deep learning era, most of the approaches \cite{scholkopf2000support,scheirer2014probability,junior2017nearest,zhang2016sparse} are based on traditional classification models such as Support Vector Machines (SVMs), Nearest Neighbors, Sparse Representation, etc. These methods usually do not scale well without careful feature engineering.
Recently, deep learning based models have shown more appealing results. The first among them is probably \cite{bendale2016towards}, which introduced Weibull-based calibration to augment the SoftMax layer of a deep network, called OpenMax. Since then, the OpenMax is further developed in \cite{rozsa2017adversarial,ge2017generative}. \cite{yoshihashi2019classification} presented the classification-reconstruction learning algorithm for open set recognition (CROSR), which utilizes latent representations for reconstruction and enables robust unknown detection without harming the known classification accuracy. \cite{oza2019c2ae} proposed the C2AE model
for open set recognition, using class conditioned auto-encoders with novel training and testing methodology. Open set recognition principles have been applied to text classification \cite{venkataram2018open,shu2017doc}, and semantic segmentation \cite{bev2019simultaneous}.

\textit{Generative methods.} Unlike discriminative models, generative approaches generate unknown samples based on Generative Adversarial Network (GAN) \cite{goodfellow2014generative} to help the classifier learn decision boundary between known and unknown samples. \cite{ge2017generative} proposed the Generative OpenMax (G-OpenMax) algorithm, which uses a conditional GAN to synthesize mixtures of known classes and finetune the closed-set classification model. G-OpenMax improves the performance of both SoftMax and OpenMax based deep network. Although G-OpenMax effectively detects unknowns in monochrome digit datasets, it fails to produce significant performance improvement on natural images. Different from G-OpenMax, \cite{neal2018open} introduced a novel dataset augmentation technique, called counterfactual image generation (OSRCI). OSRCI adopts an encoder-decoder GAN architecture to generate the synthetic open set examples which are close to knowns. They further reformulated the open set problem as classification with one additional class containing those newly generated samples. GAN-based methods also have been used to solve open set domain adaptation problem recently \cite{zhang2019improving,saito2018open}.

\textit{Out-of-distribution detection.} The open set recognition is naturally related to some other problem settings such as out-of-distribution detection \cite{vernekar2019out,joan2019likelihood,lee2017training}, outlier detection \cite{ruff2018deep}, and novelty detection \cite{perera2019ocgan}, etc. They can be incorporated in the concept of open set classification as an unknown detector. However, they do not require open-set classifiers because those models does not have discriminative power within known classes. We focus in this paper on the broader open set recognition problem.

\subsection{Flow-Based Methods}
Flow-Based (also called invertible) models have shown promises in density estimation. The original representative models are NICE \cite{dinh2014nice}, RealNVP \cite{dinh2016density} and Glow \cite{kingma2018glow}. The design ideas of these flow-based models are similar. Through the ingenious design, the inverse transformation of each layer of the model is relatively simple, and the Jacobian matrix is a triangular matrix, so the Jacobian determinant is easy to be calculated. Such models are elegant in theory, but there exists an issue in practice, \textit{i.e.}, the nonlinear transformation ability of each layer becomes weak. Apart from these flow-based models, \cite{behrmann2018invertible} proposed an Invertible Residual Network (I-ResNet), which adds some constraints to the ordinary ResNet structure to make the model invertible. The I-ResNet model still retains the basic structure of a ResNet and most of its original fitting ability. So previous experience in ResNet design can basically be re-used. Unfortunately, the density evaluation requires computing an infinite series. The choice of a fixed truncation estimator used by \cite{behrmann2018invertible} leads to substantial bias which is tightly coupled with the expressiveness of the network. It cannot be used to perform maximum likelihood because the bias is introduced in the objective and gradients. \cite{chen2019residual} improved I-ResNet, and introduced the Residual Flows, a flow-based generative model that produces an unbiased estimate of the log density. Residual Flows allows memory-efficient backpropagation through the log density computation. This allows model to use expressive architectures and train via maximum likelihood in many tasks, such as classification, density estimation and generation, etc. Our work differs from existing flow-based models in that we explicitly address a broader open-set problem, where the flow model is a sub-component.

\subsection{Flow-Based Methods for Out-of-Distribution Detection}
Flow based models have been applied to 
out-of-distribution (OOD) detection, which is relevant to open set problem. Nalisnick et al. \cite{nalisnick2019hybrid} presented a neural hybrid model created by combining deep invertible features and GLMs to filter out-of-distribution (OOD) inputs, using the model's natural ``reject" rule based on the density estimation of the flow-based component. However, this rejection rule is not guaranteed to work in all settings. The main reason is that deep generative models can assign higher likelihood to OOD inputs. Nalisnick et al. \cite{nalisnick2018deep} find that the density learned by flow-based models cannot distinguish images of common objects such as dogs, trucks, and horses (i.e. CIFAR-10) from those of house numbers (i.e. SVHN), assigning a higher likelihood to the latter when the model is trained on the former. \cite{ren2019likelihood} also observed that likelihood learned from deep generative models can be confounded by background statistics (e.g. OOD input with the same background but different semantic component). 
\cite{hendrycks2018deep} 
proposed a simple technique that uses out-of-distribution samples to teach a network heuristics to detect out-of-distribution examples, namely Outlier Exposure (OE). But this improvement is limited and sensitive to the selection of OE dataset.
\cite{kamoi2019likelihood} showed that a factor underlying this phenomenon is a mismatch between the nature of the prior distribution and that of the data distribution. They proposed the use of a mixture distribution as a prior to make likelihoods assigned by deep generative models sensitive to out-of-distribution inputs. \cite{nalisnick2019detecting} explained the phenomenon through typicality and proposed a typicality test based on batches of inputs which solves many of the failure modes.
While we also follow the same hybrid modeling direction, our work differs from \cite{nalisnick2019hybrid} in that we choose to share a common visual representation for both the classifier and the flow model and \cite{nalisnick2019hybrid} uses the output of the flow model as the input to the classifier. It is observed from our experiments that the proposed representation sharing approach is effective in our setup.

\begin{figure}[!tb]
\centering
	\includegraphics[width=1.0\textwidth]{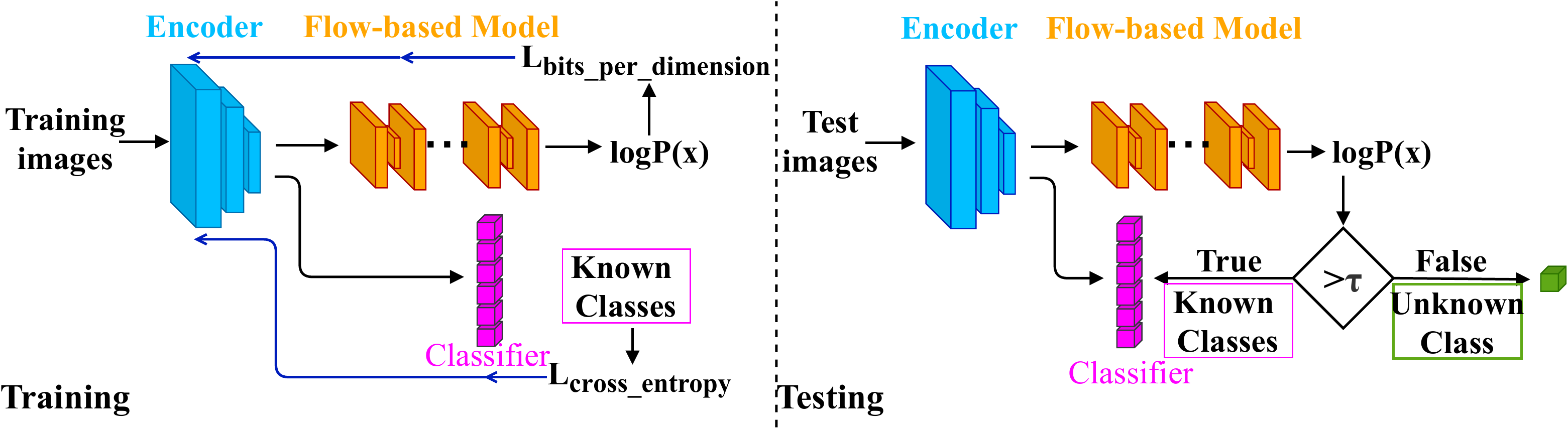}
 	\caption{ Proposed architecture for open set recognition. During the training phase (left), images are mapped into a latent feature space by the encoder, then the encoded features are fed into two branches for learning: One is typical classification learning with a classifier via cross entropy loss, and the other is density estimation with a flow-based model via its log likelihood. The whole architecture is trained in an end-to-end manner. In testing phase (right),  the $\log p(x)$ of each image is computed and then compared with the lowest $\log p(x)$ taken over the training set. If it is greater than the threshold $\tau$, it is sent to the classifier to identify its specific known class, otherwise it is rejected as an unknown sample.
 	}
	\label{method}
\end{figure}

\section{Our Approach}
We start this section by defining the open set problem and introducing the notations. Following this is an overview of our proposed approach which we call ``OpenHybrid''. After an explanation to details of each module, we introduce how to achieve open set recognition using OpenHybrid.  

\subsection{Problem Statement and Notation}
For open set recognition, given a labeled training set of instances $\mathbf{X}\in\mathbb{R}^{m\times n}$ and their corresponding labels $\mathbf{y}\in \{1, \ldots, k\}^n$ where $k$ is the number of known classes, $n$ is the total number of instances and $m$ is the dimension of each instance, we learn a model $f: \mathbf{X}\rightarrow \{1, \ldots, k+1\}^n$ such that the model accurately classify an unseen instance (in test set, not in $\mathbf{X}$) to one of the $k$ classes or an unknown class (or the ``none of the above'' class) indexed using $k+1$.

\subsection{Overview}
Figure \ref{method} overviews the training and testing procedures for the proposed method. The OpenHybrid framework consists of three modules: an encoder $\mathcal{F}$ for learning latent representations with parameters $\Theta_{f}$, a classifier $\mathcal{C}$ for classifying known classes with parameters $\Theta_{c}$, and a flow-based module $\mathcal{D}$ for density estimation with parameters $\Theta_{d}$. Existing flow-based models and their hybrid variants, which directly feed as input the original image data into the flow-based model for density estimation. Different from these works, our OpenHybrid framework directly uses the latent representation (the output of encoder $\mathcal{F}$) as the input to the flow model $\mathcal{D}$. The reason for this is that density estimation directly on the original image is susceptible to the population level background statistics (\textit{e.g.}, in MNIST, the background pixels that account for most of the image are similar), which makes it hard to detect unknown samples via exact marginal likelihood. Even in some settings with different backgrounds, unknown samples are assigned higher likelihoods than known samples, and this behavior still exists and has not been explained so far. We propose to estimate the density of latent representations instead of the original input. We find our method to be effective in all of our experimental benchmarks and we do not observe the ``higher outlier likelihood'' issue using such framework.

For classification, the classifier $\mathcal{C}$ is directly connected to the output of the encoder $\mathcal{F}$ instead of the output of the invertible transformation $\mathcal{D}$. We choose to remove the dependency of the classifier on the flow model because we believe the output of the invertible transform loses the discriminative power. We find this approach allows both the detection of unknown classes and the classification of known classes are effective.

\subsection{Training}
We define the training loss function in this section.

\vskip 0.5em
\noindent\textbf{Classification Loss.}
Given images in a batch $\{ {X}_{1}, {X}_{2}, \ldots,  {X}_{N} \}$ and their corresponding labels $\{{y}_{1}, {y}_{2}, \ldots, {y}_{N}\}$. Here $N$ is the batch size and $\forall {y}_{i} \in \{1, 2, \ldots, k\}$. Encoder $\mathcal{F}$ and classifier $\mathcal{C}$ are trained using the following cross entropy loss.
\begin{equation}\label{eq:1}
\begin{split}
\mathfrak{L}_{C} (\{\Theta_{f}, \Theta_{c}\}) = - \frac{1}{N} \sum_{i=1}^{N} \sum_{j=1}^{k} \mathbb{I}_{y_{i}} (j) \log p(y_{j}| x_{i}; \Theta_{f}, \Theta_{c})
\end{split}
\end{equation}
where $\mathbb{I}_{y_{i}}$ is an indicator function for label $y_{i}$, and $p(y_{j}| x_{i}; \Theta_{f}, \Theta_{c})$ is the probability of the $j^{th}$ class from the probability score vector predicted by $\mathcal{C}(\mathcal{F}(x_{i}))$.

\vskip 0.5em
\noindent\textbf{Density Estimation Loss.}
For unknown detection, unlike general open set methods, flow-based models directly fit the distribution of the training set, and compute the probability $p(x_{i};  \Theta_{d})$ of each training sample from the training distribution (also can be treated as the distribution of known classes) through the maximum likelihood estimation. Then, they use the model's natural reject rule based on $p(x_{i}; \Theta_{d})$ to filter unknown inputs. Although this is intuitively feasible, there are still problems as mentioned above. We suspect the problems come from the difficulty of flow models representing the original input space. So we instead estimate the density of learned latent representations $\mathcal{F}(x_{i})$. 

Flow-based model are the first key building block in our approach. These are simply high-capacity, bijective transformations with a tractable Jacobian matrix and inverse. The bijective nature of these transforms is crucial as it allows us to employ the change-of-variables formula for exact density evaluation:
\begin{equation}\label{eq:2}
\begin{split}
\log p(x_{i}; \Theta_{f}, \Theta_{d}) &= \log p(\mathcal{F}(x_i;\Theta_f);\Theta_d)\\
& =\log p(\mathcal{D}(\mathcal{F}(x_{i}; \Theta_{f}); \Theta_{d})) + \log \left| \det \frac{\partial \mathcal{D}(\mathcal{F}(x_{i}; \Theta_{f}); \Theta_{d})}{\partial\mathcal{F}(x_{i}; \Theta_{f})}\right|.
\end{split}
\end{equation}
Please note here we slightly abuse the notation for simplicity since the output of the flow model is not exactly the density of input $x$ but instead the density of its latent embedding $\mathcal{F}(x; \Theta_f)$. A simple base distribution such as a standard normal distribution is often used for $p(\mathcal{D}(\mathcal{F}(x_{i}; \Theta_{f}); \Theta_{d}))$. Tractable evaluation of Equation \ref{eq:2} allows flow-based models to be trained using the maximum likelihood with the loss function:
\begin{equation}\label{eq:3}
\begin{split}
\mathfrak{L}_{D}(\{\Theta_{f}, \Theta_{d}\})  = - \frac{1}{N} \sum_{i=1}^{N} \log p(x_{i}; \Theta_{f}, \Theta_{d})~.
\end{split}
\end{equation}
In training, we map the loss $\mathfrak{L}_{D}(\{\Theta_{f}, \Theta_{d}\})$ to bits per dimension results by normalizing the loss by the dimensionality of the flow input. In our OpenHybrid framework, there are multiple choices for the flow-based module. Considering the stability of the density estimation, we use a tractable unbiased estimate of the log density, called residual flow \cite{chen2019residual}. 

\vskip 0.5em
\noindent\textbf{Full Loss.} The complete loss function of our method is:
\begin{equation}\label{eq:4}
\begin{split}
\mathfrak{L}(\{\Theta_{f}, \Theta_{c}, \Theta_{d}\})  =\mathfrak{L}_{C} (\{\Theta_{f}, \Theta_{c}\}) + \lambda \mathfrak{L}_{D}(\{\Theta_{f}, \Theta_{d}\})
\end{split}
\end{equation}
where $\lambda$ is a scaling factor on the contribution of $p(x)$. In all of our experiments in this paper, we empirically set it to 1.

\subsection{Inference}
\textbf{Outlier Threshold.} At test time, we use the probability density estimated by flow-based module to detect unknown samples from probability distributions. This value corresponds to the probability of a sample being generated from the distribution of the training classes (known classes). Theoretically, the minimum boundary of this probability distribution in the training set is the maximum value of the outlier threshold. We assume that the known samples of the training set and the test set are from the same domain, then the outlier threshold is calculated as $
\tau = \min_{x_{i} \in \mathbf{X}}\log p(x_{i}; \Theta_{f}, \Theta_{d}) + s$, 
where $s$ is a free parameter providing slack in the margin. We estimate the outlier threshold using training samples without data augmentation.

\vskip 0.5em
\noindent\textbf{Open Set Recognition.}
Open set recognition is a classification over $k+1$ class labels, where the first $k$ labels are from the known classes the classifier $\mathcal{C}$ is trained on, and the $k+1$-st label represents the unknown class that signifies that an instance does not belong to any of the known classes. This is performed using the outlier threshold $\tau$ and the score estimated in Equation \ref{eq:2}. The outlier threshold is first calculated on training data. If the estimated probability is smaller than outlier threshold, the test instance is classified as $k+1$, which in our case corresponds to the unknown class, otherwise the appropriate class label is assigned to the instance from among the known classes. More formally, the prediction of a sample $x$ is define as
\begin{equation}\label{eq:6}
\begin{split}
pred(x) = \left\{
             \begin{array}{ll}
             k+1, &  \mathcal{D}(\mathcal{F}(x_{i}; \Theta_{f}); \Theta_{d}) < \tau, \\
             \arg\max_{j \in\{1,\ldots,k\}}\ p(y_{j} | x; \Theta_{f}, \Theta_{c}), &  \text{otherwise}~. 
             \end{array}
\right.
\end{split}
\end{equation}

\section{Experiments}
We evaluate our OpenHybrid framework and compare it with the state-of-the-art non-flow-based and flow-based open set methods. We follow other methods' protocols for fair comparisons. That is, we compare with non-flow-based open set methods without considering operating threshold while we set an unified threshold value during the comparison with flow-based methods. 

\subsection{Experiment Setups}
\textbf{Implementation.} In our experiments, the encoder, decoder, and classifier architectures are similar to those used in \cite{neal2018open}. The last layer of encoder in \cite{neal2018open} maps 512d to 100d. We moved this layer in our model to the classifier since we do not want the input dimension of flow model to be too small. So the output of our encoder is 512d instead. For flow-based model, we use the standard setup of passing the data through a logit transform \cite{dinh2016density}, followed by $10$ residual blocks. We use activation normalization \cite{kingma2018glow} before and after every residual block. Each residual connection consists of 6 layers (\textit{i.e.},
LipSwish \cite{chen2019residual} $\rightarrow$  InducedNormLinear $\rightarrow$ LipSwish $\rightarrow$ InducedNormLinear
$\rightarrow$ LipSwish $\rightarrow$ InducedNormLinear)
with hidden dimensions of 256 (the first 6 blocks) and 128 (the next 4 blocks) \cite{nalisnick2019hybrid}. We use the Adam optimizer with a learning rate 0.0001 for the encoder and flow-based module to learn log probability distribution. For training classification, we use the Stochastic Gradient Descent (SGD) with momentum 0.9 and learning rate 0.01 for TinyImageNet data, 0.1 for other data. Gradients are updated alternatively between the flow model and the classifier. The parameter $s$ is empirically set to 80. Another important factor affecting open-set performance is openness of the problem. we define the openness based on the ratio of the numbers of unique classes in training and test sets, \textit{i.e.}, $openness = 1 - \sqrt{k_\text{train}/k_\text{test}}$ where $k_\text{train}$ and $k_\text{test}$ are the number of classes in the training set and the test set, respectively.
In following experiments, we will evaluate performance over multiple openness values depending on different dataset settings.

\vskip 0.5em
\noindent\textbf{Datasets.}
We evaluate open set classification using multiple common benchmarks, such as MNIST \cite{lecun2010mnist}, SVHN \cite{netzer2011reading}, CIFAR10 \cite{krizhevsky2009learning}, CIFAR+10, CIFAR+50 and TinyImageNet \cite{le2015tiny} datasets. We reuse the data splits provided by \cite{neal2018open}. 
\begin{itemize}
\item \textit{MNIST, SVHN, CIFAR10}:  All three datasets contain 10 categories. MNIST are monochrome images with hand-written digits, and it has 60k 28$\times $28 gray images for training and 10k for testing. SVHN are street view house numbers, consisting of ten digit classes each with between 9981 and 11379 32$\times $32 color images. To validate our method on non-digital images, we apply the CIFAR10 dataset, which has 50k 32$\times $32 natural color images for training and 10k for testing. Each dataset is partitioned at random into 6 known and 4 unknown classes. In these settings, the openness score is fixed to 22.54\%.

\item \textit{CIFAR+10, CIFAR+50}: To test the method in a range of greater openness scores, we perform CIFAR+$U$ experiments using CIFAR10 and CIFAR100 \cite{krizhevsky2009learning}. 4 known classes are sampled from CIFAR10 and $U$ unknown classes are drawn randomly from the more diverse CIFAR100 dataset. Openness scores of CIFAR+10 and CIFAR+50 are 46.54\% and 72.78\%, respectively. 

\item \textit{TinyImageNet}: For the larger TinyImagenet dataset, which is a 200-class subset of ImageNet, we randomly sampled 20 classes as known and the remaining classes as unknown. In this setting, the openness score is 68.37\%.

\end{itemize}

The out-of-distribution (OOD) detection community often evaluates methods on cross-dataset setups \cite{hendrycks2018deep,nalisnick2019hybrid,nalisnick2019detecting,ren2019likelihood}, such as training on CIFAR10 and testing on CIFAR100. So we perform extra experiments on two such settings  between CIFAR10 and CIFAR100 and report results comparable to OOD literature.

\vskip 0.5em
\noindent\textbf{Metrics.}
Open set classification performance can be characterized by F-score or AUROC (Area Under ROC Curve) \cite{geng2018recent}. AUROC is commonly reported by both open set recognition and out-of-distribution detection literature. So we mainly use AUROC to compare with existing methods. We adopt F-score in some of our experiments as it also measures the in-distribution classification performance. For both metrics, higher values are better.

\subsection{Results}
\textbf{Comparison with Non-flow-based Methods.}
We compare OpenHybrid against the following non-flow-based baselines:
\begin{enumerate}
    \item {\textit{SoftMax}}: A standard confidence-based method for
    open-set recognition by using SoftMax score of a predicted class. 
    \item {\textit{OpenMax}} \cite{bendale2016towards}: This approach augments the baseline classifier with a new OpenMax layer replacing the SoftMax at the final layer of the network.
    \item {\textit{G-OpenMax}} \cite{ge2017generative}: A direct extension of OpenMax method, which trains networks with synthesized unknown data by using a Conditional GAN. 
    \item {\textit{OSRCI}} \cite{neal2018open}: An improved version of G-OpenMax work, which uses a specific data augmentation technique called counterfactual image generation to train the classifier for the $k+1$-st class. 
    \item {\textit{C2AE}} \cite{oza2019c2ae}: This approach uses class conditioned auto-encoders with novel training and testing methodologies for open set recognition.
    \item {\textit{CROSR}} \cite{yoshihashi2019classification}: A deep open set classifier augmented by latent representation learning which jointly classifies and reconstructs the input data. 
\end{enumerate}

\begin{table}[!t]
\setlength\tabcolsep{5pt}
\begin{center}
\caption{AUROC for comparisons of our method with recent open set methods. Results averaged over 5 random class partitions. The best results are highlighted in \textbf{bold}.
\label{table-1}}
\vskip -1em
\resizebox{\linewidth}{!}{
\begin{tabular}{ccccccc}
\toprule
\bf Method          & \bf MNIST & \bf SVHN & \bf CIFAR10 & \bf CIFAR+10 & \bf CIFAR+50 & \bf TinyImageNet \\ \midrule
SoftMax         &   0.978    &  0.886    &    0.677     &     0.816     &    0.805      &      0.577        \\
OpenMax \cite{bendale2016towards}        &    0.981   &  0.894    &    0.695     &     0.817     &     0.796     &    0.576          \\
G-OpenMax \cite{ge2017generative}      &   0.984    &   0.896   &    0.675     &    0.827      &    0.819       &    0.580          \\
OSRCI \cite{neal2018open}          &   0.988    &   0.910    &     0.699    &    0.838      &     0.827     &     0.586         \\
C2AE \cite{oza2019c2ae}           &   0.989    &   0.922   &    0.895     &     0.955     &    0.937      &      0.748        \\
CROSR \cite{yoshihashi2019classification}          &   0.991    &   0.899   &     0.883    &     0.912     &       0.905   &       0.589       \\ \midrule
OpenHybrid (ours) &   \textbf{0.995}    &  \textbf{0.947}    &     \textbf{0.950}     &       \textbf{0.962}    &      \textbf{0.955}     &        \textbf{0.793}       \\ \bottomrule
\end{tabular}}
\end{center}
\end{table}


Table \ref{table-1} presents the open set recognition performance of our method and non-flow-based baselines on six datasets. Our approach OpenHybrid outperforms all of the baseline methods, which demonstrates the effectiveness of our approach. It is interesting to note that our method on MNIST dataset produces a minor improvement compared to the other methods. The main reason is that the MNIST is relatively simple, and the results of all methods on it are almost saturated. But for other relatively complex databases, our method performs significantly better than the the baseline methods, especially for natural images, such as CIFAR (6\% better than the second best) and TinyImageNet (5\% better than the second best).

\vskip 0.5em
\noindent\textbf{Comparison with Flow-based Methods.}
We compare our approach against our implementations of the following flow-based approaches:
\begin{enumerate}
    \item {\textit{DIGLM} \cite{nalisnick2019hybrid}}: A neural hybrid model consisting of a linear model defined on a set of features computed by a deep invertible transformation. It uses the model’s natural reject rule based on the generative component $p(x)$ to detect unknown inputs. The threshold is setted as $\min_{\mathbf{x}\in \mathbf{X}} p(x; \theta ) - c$, where the minimum is taken over the training set and $c$ is a free parameter providing slack in the margin. 
    \item {\textit{OE}} \cite{hendrycks2018deep}: A training method leveraging an auxiliary dataset of unknown samples to improve unknown detection. The framework is the same as DIGLM, except that during training, a margin ranking loss on the log probabilities of training and outlier exposure samples is used to update the flow-based model. In this experiment, we use counterfactual images generated by \cite{neal2018open} from training samples as its outlier exposure dataset.
\end{enumerate}

\begin{table}[!t]
\begin{center}
\caption{ AUROC for our methods and flow-based baselines. Results are averaged over 5 random class partitions. The best results are highlighted in \textbf{bold}.}
\label{table-2}
\setlength\tabcolsep{5pt}
\vskip -1em
\resizebox{\linewidth}{!}{
\begin{tabular}{ccccccc}
\toprule
\bf Method          & \bf MNIST & \bf SVHN & \bf CIFAR10 & \bf CIFAR+10 & \bf CIFAR+50 & \bf TinyImageNet  \\ \toprule
DIGLM     &       0.643   &  0.559    &    0.583    &   0.590    &    0.594     &     0.520      \\
DIGLM + OE      &   0.721   &  0.643    &    0.655    &   0.670     &    0.671    &      0.596       \\ 

OpenHybrid (ours) &   \textbf{0.995}    &  \textbf{0.947}    &     \textbf{0.950}     &       \textbf{0.962}    &      \textbf{0.955}     &        \textbf{0.793}         \\ \bottomrule
\end{tabular}}
\end{center}
\end{table}

Table \ref{table-2} shows the AUROC of our method and the flow-based baselines in different datasets. We observe that our method consistently outperforms the baseline methods significantly under all open set benchmarks. The same trend is observed for the f-score metric, \text{e.g.}, we achieved 0.865 in CIFAR10 while DIGLM achieves only 0.673 and DIGLM+OE achieves 0.701).

\vskip 0.5em
\noindent\textbf{Cross-dataset OOD settings.} We further evaluate our approach on two cross-dataset settings: training on CIFAR10 and testing on CIFAR100 and vice versa. We compare the AUROC of our method directly with the numbers reported in \cite{hendrycks2018deep}. The results suggest that our approach is still competitive in such settings. It is worth noting that training on CIFAR100 and testing on CIFAR10 is a harder task, probably due to the higher number of training classes. Our approach achieves higher gains ($+10\%$) in this setting.


\subsection{Discussion}
\textbf{The benefit of joint training.}
We further compare the end-to-end trained OpenHybrid with a different training strategy based on alternative training. The framework is still the same. However, during training, the encoder and classifier are pretrained first on the training data. The flow-based model was then trained separately with both encoder and classifier being frozen. Table \ref{tab:ablation} shows a comparison between the two methods using F-score. The slack parameter $s$ is chosen to be 80 for all datasets. We observe that joint training consistently outperforms OpenHybrid with a fixed pretrained encoder.

\begin{table}[t]
\centering

\begin{minipage}{.45\linewidth}
\caption{AUROC for cross-dataset out-of-distribution detection between CIFAR-10 and CIFAR-100.}
\label{tab:cifar-cross}
\setlength\tabcolsep{7pt}
\vskip -0.3em
\resizebox{\linewidth}{!}{
\begin{tabular}{lcc}
\toprule
\bf  Train $\rightarrow$ Test (OOD)         & \bf OE \cite{hendrycks2018deep} & \bf Ours    \\ \toprule

 CIFAR10 $\rightarrow$ CIFAR100          &   0.933    &   \bf 0.951   \\
 CIFAR100 $\rightarrow$ CIFAR10 &   0.757    &  \textbf{0.856}      \\
\bottomrule
\end{tabular}}
\end{minipage}
\quad
\begin{minipage}{.45\linewidth}
\caption{F-scores\protect\footnotemark\  of the proposed OpenHybrid models using pretrained encoder and joint training.}
\label{tab:ablation}
\setlength\tabcolsep{2pt}
\vskip -0.5em
\resizebox{\linewidth}{!}{
\begin{tabular}{ccccc}
\toprule
\bf Method          & \bf MNIST & \bf SVHN & \bf CIFAR10  \\ \toprule

Pretrained encoder          &   0.847    &   0.842   &   0.791   \\
Joint training &   \textbf{0.942}    &  \textbf{0.912}    &     \textbf{0.865}      \\ \bottomrule
\end{tabular}}
\end{minipage}
\end{table}

\footnotetext{\footnotesize For readers who are interested in classification accuracy: Our approach achieves overall accuracy 0.947 in MNIST, 0.929 in SVHN and 0.868 in CIFAR10. However, we believe F-score is a better measurement which considers data imbalance.}

\vskip 0.5em
\noindent\textbf{A study on the parameters.} Our loss function contains a trade-off parameter $\lambda$. We varied this value among 0.5, 1 and 2 in the MNIST dataset and observed AUROC scores 0.993, 0.995, and 0.998, respectively. The model seems not sensitive to this variable but it is a parameter that can be tuned to further improve performance. Another important parameter is the number of residual blocks in the flow model. We varied this value among 4, 8, 10 and 16 in CIFAR10 benchmark. Surprisingly, we still observe stable AUROC results (0.945, 0.948, 0.950, 0.958). So, for practitioners who may have resource constraints, it is advised to consider a smaller flow-based network when using the OpenHybrid framework.

\begin{figure}[!tb]
\centering
	\includegraphics[width=0.55\linewidth]{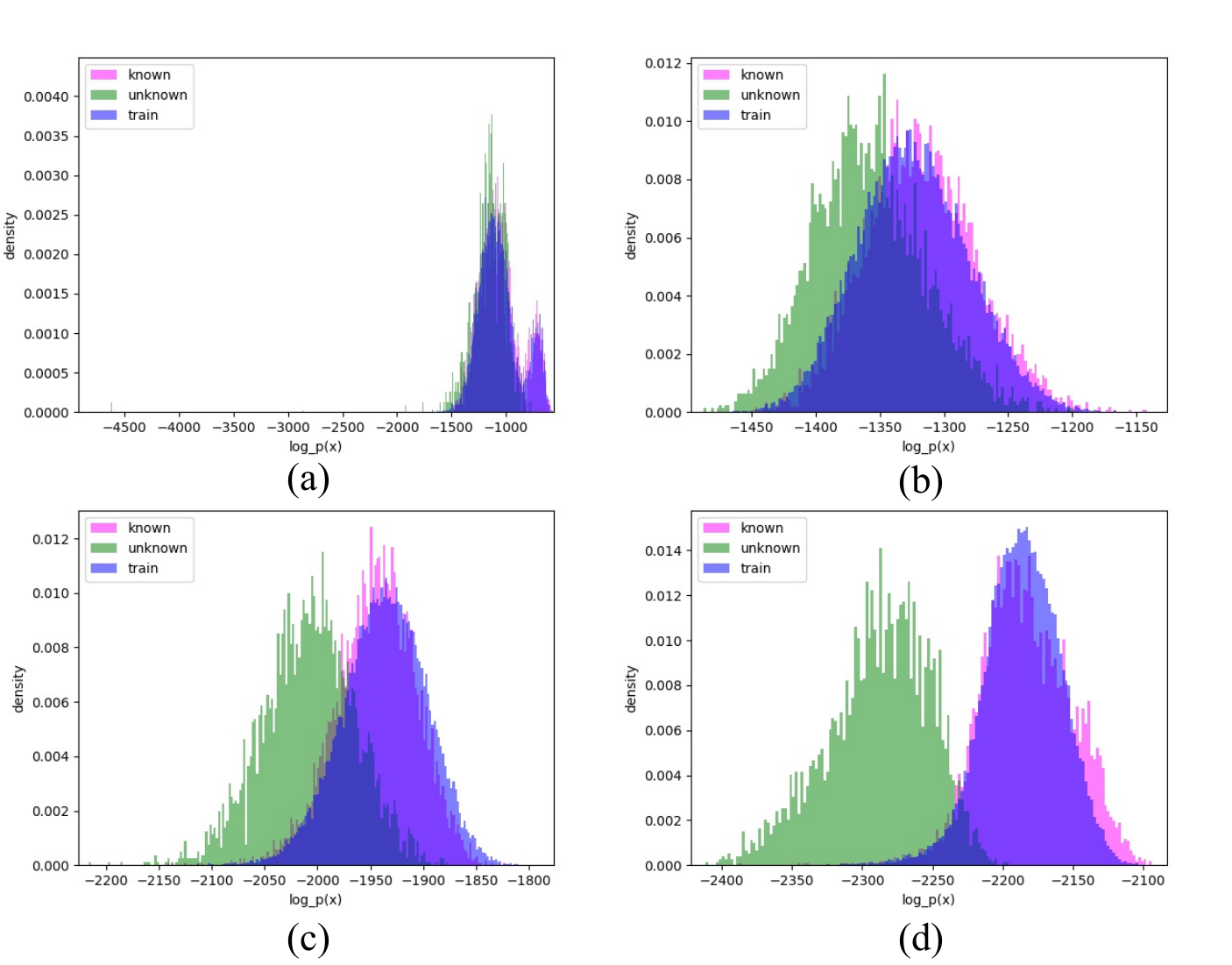}
	\includegraphics[width=0.42\linewidth]{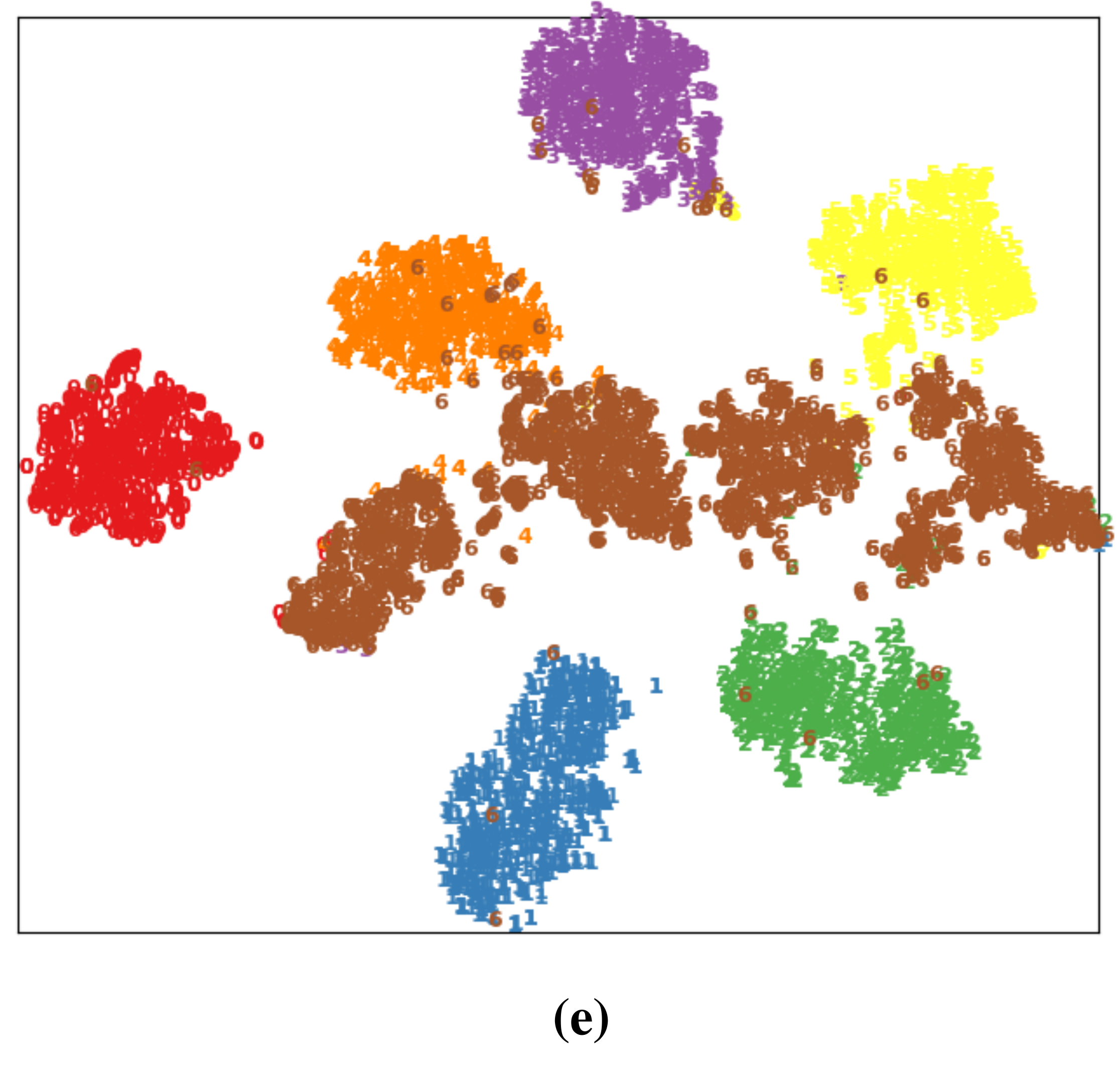}
 	\caption{ Left: Histograms of log-likelihoods for MNIST (0-5 as known classes and 6-9 as unknown classes) made by (a) DIGLM, (b) DIGLM + OE, (c) OpenHybrid with pretrained encoder and end-to-end OpenHybrid. The blue color indicates training samples, the pink indicates known samples in the test, and the green is unknown samples. Right (e): t-SNE visualization of the latent space by end-to-end OpenHybrid. Different colors represent different classes. Brown color represents the unkown digits (6-9).
 	}
	\label{MNIST}
\end{figure}

\vskip 0.5em
\noindent\textbf{A visualization of the estimated density.} Figure \ref{MNIST} (left) shows the histograms of log-likelihoods for MNIST (0-5 as known classes and 6-9 as unknown classes) made by DIGLM, OE, OpenHybrid with pretrained encoder and OpenHybrid with joint training. For DIGLM (a), the three histograms almost overlap so it is impossible to detect the unknown class by setting a threshold. The density estimation is improved with the help of OE (b), however, there is still a large area of overlap. The distribution overlap becomes further smaller but still not ideal when using OpenHybrid with pretrained encoder (c). In contrast, we observed the end-to-end OpenHybrid (d) produces the histogram of unknown samples well separated from those of known samples.  

\vskip 0.5em
\noindent\textbf{A visualization of the latent space.} Figure \ref{MNIST}(e) shows a t-SNE \cite{tsne} plot of the latent space learned by end-to-end OpenHybrid. The brown color represents the unkonwn classes (digit 6,7,8,9) which is well separated from other color (known classes from 0 to 5). Interestingly, the model also learns to separate digits 6-9, which is in an unsupervised fashion. Although the MNIST dataset is simple compared to other real datasets, this result shows the potential of representation learning using hybrid models as a promising research direction.

\begin{figure}[!tb]
\centering
	\includegraphics[width=0.9\textwidth]{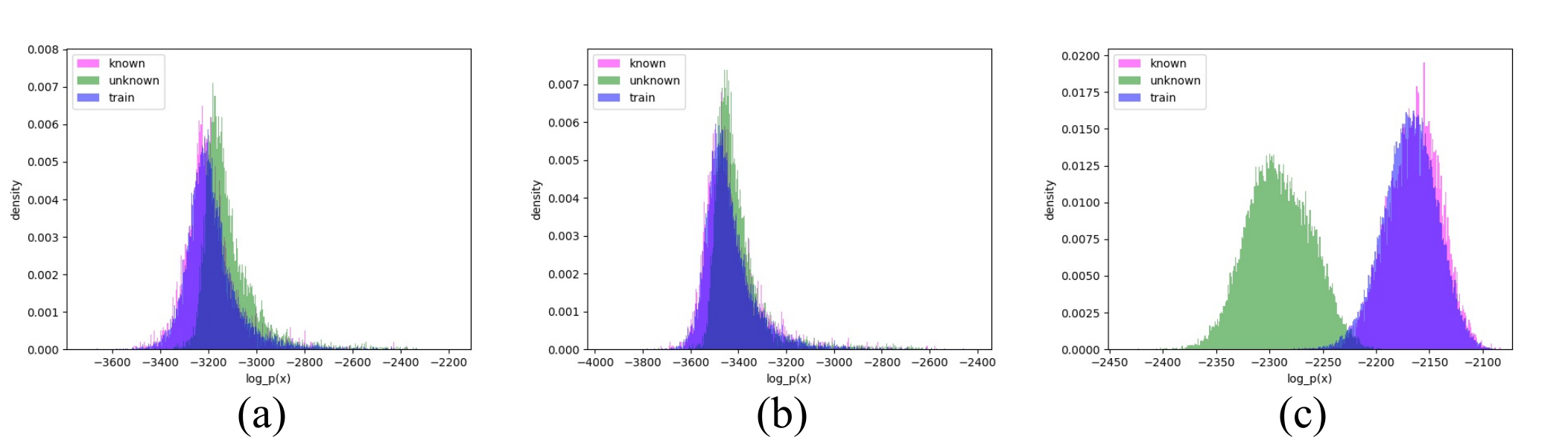}
 	\caption{ Histograms of log-likelihoods for CIFAR10 (known samples) and SVHN (unknown samples) made by (a) DIGLM, (b) DIGLM + OE and (c) the proposed OpenHybrid. The blue color indicates training samples, the pink indicates known samples in the test, and the green is unknown samples.
 	}
	\label{CIFAR_SVHN}
\end{figure}

\vskip 0.5em
\noindent\textbf{A disappeared issue of flow-based models.} Nalisnick et al.  \cite{nalisnick2018deep} raised the issue that the flow-based model trained on CIFAR10 will assign a higher log-likelihood value to SVHN. So we further conduct an experiment on this setting, where we use the full 10 classes of the CIFAR10 as known classes, and the SVHN as an unknown class. Our approach achieves 0.998 AUROC on this setting. Figure \ref{CIFAR_SVHN} shows the histograms of log-likelihoods under this setting. Similar to the observation made by \cite{nalisnick2018deep}, in Figure \ref{CIFAR_SVHN}(a), the histogram of unknown samples (green) is shifted more to the right than that of known samples (blue and pink), \textit{i.e.}, unknown samples are assigned a larger log-likelihood value than known samples. In Figure \ref{CIFAR_SVHN}(b), OE seems to help but it does not fully address the problem as well. Our method is shown in Figure \ref{CIFAR_SVHN}(c) which clearly distinguish the two distributions. The histogram of unknown samples is almost entirely to the left of known samples. We believe a potential reason is that the original input space is a multimodal distribution and our method projects the input data into a latent space which is probably more suitable to the unimodal assumption of flow-based models. While we are unable to prove this theoretically, we hope our results could inspire future works on deeper understanding of flow-based models.

\section{Conclusion}

We presented the OpenHybrid framework for open set recognition. Our approach is built upon a flow-based model for density estimation and a discriminative classifier, with a shared latent space. Extensive experiments show that our approach achieves the state of the art. A common issue of flow-based models is that they often assign larger likelihood to out-of-distribution samples. We empirically observe on various datasets that this issue disappear by learning a joint feature space. Ablation study also suggests that joint training is another key contributing factor to the superior open set recognition performance.

\subsubsection{Acknowledgement.}
We would like to thank Balaji Lakshminarayanan and Olaf Ronneberger for meaningful discussions. This research was supported by the National Science Foundation of China under Grants 61772257 and the Fundamental Research Funds for the Central Universities 020914380080.

\clearpage
%
%
\bibliographystyle{splncs04}
\bibliography{egbib}
\end{document}